# Predicting the Success of Domain Adaptation in Text Similarity

**Nicolai Pogrebnyakov**[*†]   **Shohreh Shaghaghian**[*]

[*]Thomson Reuters Labs, Canada    [†]Copenhagen Business School, Denmark
Emails: firstname.lastname@thomsonreuters.com

## Abstract

Transfer learning methods, and in particular domain adaptation, help exploit labeled data in one domain to improve the performance of a certain task in another domain. However, it is still not clear what factors affect the success of domain adaptation. This paper models adaptation success and selection of the most suitable source domains among several candidates in text similarity. We use descriptive domain information and cross-domain similarity metrics as predictive features. While mostly positive, the results also point to some domains where adaptation success was difficult to predict.

## 1 Introduction

Since the data-hungry deep learning models have beaten state-of-the-art performances in different natural language processing (NLP) tasks, many efforts have been made to deal with the scarcity of labeled data (Wang et al., 2020; Settles, 2010; Kouw and Loog, 2019). One of the main avenues taken by researchers of this field is investigating the portability of models between different data distributions, often referred to as different domains (Luo et al., 2019; Gururangan et al., 2020). While multiple approaches have been proposed to make this portability feasible and efficient, it is still unclear how to predict the adaptability of two domains in advance. It is particularly important to address this gap because almost all domain adaptation approaches adjust a model to a new domain at the expense of more computational resources. Therefore, in practice it is neither desirable nor scalable to try all possible dataset candidates.

Most relevant existing work seeks to identify key factors that can be used to justify why transfer learning between two domains work (Asch and

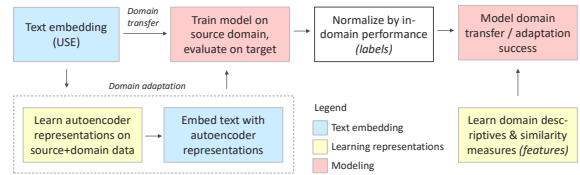

Figure 1: The process of modeling the success of domain transfer & adaptation.

Daelemans, 2010; Dai et al., 2019; Kashyap et al., 2021; Mou et al., 2016; Shah et al., 2018). In practice, however, one needs to be able to quantitatively select a set of existing datasets that can best be adapted to a certain domain for a certain task.

We propose a simple yet effective approach to predict the success of transfer or adaptation, with the hope of drawing the research community's attention to this gap. We use the term *domain transfer* (DT) when a model trained on one domain is simply used for inference in another domain. *Domain adaptation* (DA) is used for approaches that bridge the source and target domain representations (e.g., by mapping or aligning feature spaces of the two domains) so that a model trained on labeled source data (and unlabeled target data) performs well in the target domain. While for the experiments in this paper we focus on the task of text similarity and autoencoder approaches to DA, the proposed process, shown in Figure 1, can be easily applied to other NLP tasks and other unsupervised DA approaches.

**Domain Adaptation.** The need for domain adaptation arises when a model trained using labeled data from one (source) domain needs to be applied to another (target) domain with a different data distribution (Miller, 2019). We focus specifically on unsupervised domain adaptation, which learns using unlabeled data in both source and target domains. Model-based approaches to unsupervised DA have been classified into modifying the feature space and augmenting the loss function (see Ramponi and Plank (2020) for a comprehensive review).

**Domain Similarity.** Extant studies have proposed a variety of measures to quantitatively express similarity between a pair of domains. Dai et al. (2019) define three main metrics to measure different aspects of similarity between source and target datasets and investigate how these measures correlate with the effectiveness of named entity recognition tasks. Target vocabulary coverage, language model perplexity, and word vector variance are used as these similarity measures. Asch and Daelemans (2010) show a correlation between six similarity metrics based on word frequency, and the performance of some part-of-speech tagging tasks.

**Autoencoders for Domain Adaptation.** Stacked denoising autoencoders (SDA) learn latent representations that align feature spaces of the source and target domains (Ramponi and Plank, 2020; Vincent et al., 2010). SDA first add noise to input, such as dropout or Gaussian noise, and then aim to reconstruct the uncorrupted input (Gondara, 2016).

A further development of this approach is marginalized SDA, which marginalizes the reconstruction loss. The solution to the loss has a closed form, which lowers computational cost and improves scalability compared to the original SDA (Ramponi and Plank, 2020; Chen et al., 2012).

Inspired by another approach to DA, domain adversarial (Ganin et al., 2016), Clinchant et al. (2016) add a regularization term based on a domain classifier to the reconstruction loss. We refer to this approach as marginalized SDA with domain regularization (mSDAR). There also exists a closed-form solution to that loss, and that approach was shown to outperform marginalized SDA.

## 2 Data

We use 11 publicly available semantic text similarity datasets. Seven of them were obtained from StackExchange forums, with data from 2015 to November 2020: Apple, AskUbuntu, Math, StackOverflow, Stats, SuperUser and Unix[1]. We also use Quora Question Pairs, Microsoft Research Paraphrase Corpus (MRPC) (Dolan and Brockett, 2005), Paraphrase Adversaries from Word Scrambling (PAWS) (Zhang et al., 2019) and Sentences Involving Compositional Knowledge—Relatedness (SICK) (SemEval, 2014). All datasets contain binary labels indicating whether the text pair is similar or not. The exception is SICK, where we convert the original relevance score of 1—5 into a binary score of 0 (not semantically similar) if the relevance score is below 4, and 1 otherwise. Each of the 11 datasets is considered a separate domain.

## 3 Modeling Domain Transfer and Adaptation

Figure 1 shows the process we use to implement DT and DA and train a model that can best identify a proper source domain for a particular target domain. We start with embedding text in each of the 11 domains with the Universal Sentence Encoder (USE) (Cer et al., 2018). For DT, USE representations are used directly to train models in a source domain $S$ and evaluate the performance in the target domain $T$. For DA, we implement both SDA and mSDAR. A three-layer SDA is trained on each source-target domain pair. Text from the source and target domains is embedded with USE and corrupted with Gaussian noise, whose parameters are estimated from the hidden representation of the previous layer. In mSDAR, the hyperparameters we use are 5 layers, the target regularization parameter $\lambda = 1$, dropout probability 0.6 and the regularization objective $\mathbf{R} = \mathbf{1}$. These parameters have the same meaning as in Clinchant et al. (2016). Both SDA and mSDAR encoders are then used to embed the USE representations of the text in the source and target domains. Figure 2 shows the original and mSDAR representations for StackOverflow and SuperUser domains, demonstrating the effect of mSDAR on aligning the feature spaces of the two domains.

The representations described above are used to train a dense 3-layer neural network in the source domain and evaluate its performance in the target domain by reporting the F1-score. (We train three such models for each domain pair and average their performance.) We denote this cross-domain performance by $F1_{ST}$. In order to make the DT and DA results robust to the relative difficulty of learning in different domains, we normalize $F1_{ST}$ by the in-domain F1-score, $F1_{TT}$, which denotes the performance of the fully supervised model trained and evaluated in the same domain. The normalized F1-score, averaged over all domain pairs, is 0.775 for DT, 0.799 for SDA and 0.817 for mSDAR. This is in line with previous work showing better performance of mSDAR over SDA (Clinchant et al., 2016).

---
[1] Obtained from https://data.stackexchange.com

## 4 Domain Similarity Measures

Considering a source domain $S$ and a target domain $T$ with unigram sets $U_S$ and $U_T$, we define a set of features $F^{ST} = \{f_1^{ST}, \ldots, f_{10}^{ST}\}$ as follows.

**Unigram Coverage**. The simplest metric to evaluate the similarity of two domains is the percentage of their common unigrams. We use the ratio of common unigrams in the source $f_1^{ST} = \frac{|U_S \cap U_T|}{|U_S|}$ and target $f_2^{ST} = \frac{|U_S \cap U_T|}{|U_T|}$ domains as two features for the classifiers.

**Dataset Size.** The number of labeled data points in source ($f_3^{ST}$) and target ($f_4^{ST}$) domains as well as the average number of tokens per example for the source and target domains ($f_5^{ST}$ and $f_6^{ST}$) are additional features we use for the classifiers.

**Distribution Similarity**. In order to measure the similarity of how tokens have been distributed in the two domains, we add Rényi divergence ($f_7^{ST}$) (Asch and Daelemans, 2010) and KL divergence ($f_8^{ST}$) (Plank and van Noord, 2011) to the set of features. We use α = 0.99 as the value of the parameter in Rényi divergence.

**Language Similarity**. Similar to Dai et al. (2019), we train a trigram language model in each domain and evaluate its perplexity on other domains ($f_9^{ST}$). Since the target domain is expected to have many trigrams that are not seen in the source domain, we apply Kneser-Ney smoothing to account for those unseen trigrams (Kneser and Ney, 1995). We also use word vector variance between the source and target domains ($f_{10}^{ST}$) (Dai et al., 2019). This variance is calculated as $\frac{1}{|U_S \cap U_T| d} \Sigma_{v \in U_S \cap U_T} \Sigma_{j=1}^{d} \left| W_{Sv}^j - W_{Tv}^j \right|$, where $W_{Sv}^j$ and $W_{Tv}^j$ are the $j^{th}$ element of the vector for word v respectively in the source and target domains. We use Word2vec Skipgram with vector length of 300.

## 5 Source Domain Selection

**Success Prediction and Order Ranking**. We evaluate two different approaches to select one or multiple source domains for a particular target domain. In the first approach, we train a classifier to predict if a domain can be a good candidate for transfer or adaptation to a specific target domain.

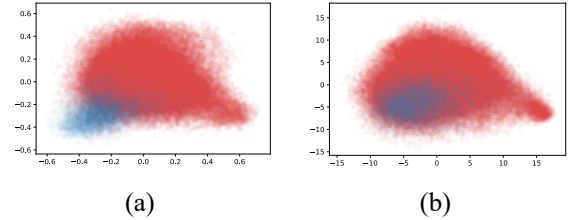

Figure 2. Original (a) and mSDAR (b) representations of text in the StackOverflow (red) and SuperUser (blue) domains (PCA projection).

We consider transfer or adaptation successful if the ratio $\frac{F1_{ST}}{F1_{TT}}$ is greater than 80% i.e., if it can achieve at least 80% of the performance of a fully supervised model on the target domain. We refer to the classifier trained in this approach as *Success Predictor*. In the second approach, irrespective of what percentage of a fully supervised model performance can be achieved, we order the existing source domains for each target domain. Therefore, we model the problem as a ranking problem and refer to the trained model as *Domain Ranker*. This ranking problem can be modeled as a binary classifier, in which a sample corresponds to the performance of two source domains $S_1$ and $S_2$ for a specific target domain $T$. The label is one if $F1_{S_1T} \geq F1_{S_2T}$ and zero otherwise.

**Performance Evaluation**. While the F1-scores and accuracies reflect how well the trained classifiers work, the original purpose of defining these two approaches was to find the *best candidates* for source domains. Hence, we also show the performance of the two approaches based on ordering-based metrics. To find the orderings for each target domain by Success Predictor, we order the source domains based on the predicted probability of the binary classifier. In Domain Ranker, we sort the source pairs using the pairwise preference predicted by the classifier. We handle the inconsistencies caused by incorrect predictions using the multi-sort algorithm proposed by Maystre and Grossglauser (2017).

To train the Success Predictor and Domain Ranker models, we use a set of features $F^{ST}$ described in section 4. For both approaches, we train a binary XGBoost classifier with 5-fold cross-validation.

## 6 Results

**Modeling Domain Adaptation**

Ideally, the best way to evaluate the performance of the two approaches is to train the model on some set of domains and test it on orderings of an entirely different set of domains. However, since we only have 11 domains, to make the most use out of this small data, we train the classifier on multiple train-test splits and report the performance metrics of the trained binary classifier each time.

|  | | DT ||||||  SDA |||||| mSDAR ||||||
|---|---|---|---|---|---|---|---|---|---|---|---|---|---|---|---|---|---|---|---|
|  | Target | F1 | Acc | CRP | Top1 | Top3 | Top5 | F1 | Acc | CRP | Top1 | Top3 | Top5 | F1 | Acc | CRP | Top1 | Top3 | Top5 |
| **Success Predictor** | Apple | 1 | 1 | 0.4 | 1 | 0.67 | 1 | 0.89 | 0.9 | 0.3 | 0 | 0.67 | 1 | 0.89 | 0.9 | 0.4 | 1 | 1 | 1 |
|  | AskUbuntu | 1 | 1 | 0.3 | 0 | 0.67 | 1 | 0.86 | 0.9 | 0 | 0 | 0.67 | 0.8 | 1 | 1 | 0.2 | 0 | 0.67 | 0.8 |
|  | MRPC | 0.71 | 0.6 | 0.2 | 0 | 0.67 | 0.6 | 0.93 | 0.9 | 0.2 | 0 | 0.33 | 0.6 | 0.93 | 0.9 | 0.3 | 0 | 0.67 | 0.8 |
|  | Math | 0.67 | 0.6 | 0.2 | 0 | 0.67 | 0.6 | 0.6 | 0.6 | 0 | 0 | 0.33 | 0.8 | 0.86 | 0.8 | 0.1 | 0 | 0.33 | 0.6 |
|  | PAWS | 0 | 0 | 0.3 | 0 | 0.33 | 0.8 | 0.18 | 0.1 | 0.3 | 0 | 0.33 | 0.6 | 0.18 | 0.1 | 0 | 0 | 0.33 | 0.6 |
|  | Quora | 0 | 0 | 0.3 | 0 | 0.67 | 1 | 0 | 0 | 0.3 | 1 | 0.67 | 0.8 | 0 | 0.1 | 0.3 | 1 | 0.33 | 0.8 |
|  | SICK | 0.89 | 0.8 | 0.5 | 0 | 0.67 | 0.8 | 0.88 | 0.8 | 0.2 | 0 | 0 | 0.4 | 0.93 | 0.9 | 0.3 | 0 | 0.33 | 1 |
|  | StackOverflow | 0.67 | 0.8 | 0.1 | 0 | 0.67 | 0.6 | 0 | 0.8 | 0.3 | 0 | 1 | 0.6 | 0.75 | 0.8 | 0.1 | 0 | 1 | 0.6 |
|  | Stats | 0.67 | 0.8 | 0.2 | 1 | 0.33 | 0.8 | 0.67 | 0.9 | 0.5 | 1 | 0.67 | 0.6 | 0.4 | 0.7 | 0.4 | 1 | 0.67 | 0.8 |
|  | SuperUser | 0.89 | 0.9 | 0.2 | 0 | 0.67 | 1 | 1 | 1 | 0.1 | 0 | 0.67 | 0.8 | 0.86 | 0.9 | 0.3 | 1 | 1 | 0.8 |
|  | Unix | 1 | 1 | 0.3 | 1 | 0.67 | 1 | 1 | 1 | 0.7 | 1 | 1 | 1 | 0.86 | 0.9 | 0.5 | 1 | 1 | 1 |
|  | **AVERAGE** | **0.68** | **0.68** | **0.27** | **0.27** | **0.61** | **0.84** | **0.64** | **0.72** | **0.26** | **0.27** | **0.58** | **0.73** | **0.70** | **0.73** | **0.26** | **0.45** | **0.67** | **0.78** |
| **Domain Ranker** | Apple | 0.86 | 0.89 | 0.8 | 1 | 1 | 1 | 0.97 | 0.98 | 0.8 | 1 | 1 | 1 | 0.89 | 0.91 | 0.5 | 1 | 0.67 | 1 |
|  | AskUbuntu | 0.85 | 0.89 | 0.7 | 0 | 0.67 | 1 | 0.86 | 0.91 | 0.4 | 0 | 0.67 | 1 | 0.77 | 0.84 | 0.5 | 0 | 0.67 | 0.8 |
|  | MRPC | 0.77 | 0.78 | 0.2 | 1 | 0.67 | 0.8 | 0.65 | 0.76 | 0.2 | 0 | 0.67 | 1 | 0.72 | 0.84 | 0.3 | 1 | 0.67 | 1 |
|  | Math | 0.73 | 0.76 | 0.1 | 0 | 0.67 | 0.6 | 0.67 | 0.71 | 0.1 | 0 | 0.33 | 0.6 | 0.65 | 0.76 | 0.2 | 0 | 0.33 | 0.8 |
|  | PAWS | 0.44 | 0.56 | 0.2 | 0 | 0.33 | 0.6 | 0.69 | 0.76 | 0.2 | 0 | 0.67 | 1 | 0.59 | 0.76 | 0.1 | 0 | 0.33 | 0.8 |
|  | Quora | 0.86 | 0.84 | 0.4 | 0 | 0.67 | 1 | 0.87 | 0.87 | 0.3 | 0 | 0.67 | 1 | 0.76 | 0.82 | 0.2 | 0 | 0.67 | 0.8 |
|  | SICK | 0.87 | 0.87 | 0.2 | 0 | 0.33 | 1 | 0.54 | 0.62 | 0.2 | 0 | 0.33 | 0.6 | 0.76 | 0.84 | 0.4 | 0 | 0.67 | 1 |
|  | StackOverflow | 0.88 | 0.89 | 0.5 | 0 | 0.67 | 1 | 0.78 | 0.8 | 0.5 | 0 | 0.33 | 0.8 | 0.86 | 0.87 | 0.5 | 1 | 0.67 | 1 |
|  | Stats | 0.91 | 0.91 | 0.7 | 1 | 0.67 | 0.8 | 0.86 | 0.87 | 0.4 | 1 | 0.67 | 0.8 | 0.83 | 0.87 | 0.4 | 1 | 0.67 | 0.8 |
|  | SuperUser | 0.94 | 0.93 | 0.5 | 0 | 1 | 1 | 0.94 | 0.93 | 0.5 | 0 | 1 | 0.8 | 0.9 | 0.91 | 0.6 | 1 | 1 | 0.8 |
|  | Unix | 0.92 | 0.91 | 0.4 | 0 | 0.67 | 1 | 0.89 | 0.89 | 0.3 | 1 | 0.67 | 0.8 | 0.93 | 0.93 | 0.5 | 0 | 1 | 0.8 |
|  | **AVERAGE** | **0.82** | **0.84** | **0.43** | **0.27** | **0.67** | **0.89** | **0.79** | **0.83** | **0.35** | **0.27** | **0.64** | **0.84** | **0.79** | **0.85** | **0.38** | **0.45** | **0.67** | **0.87** |

Table 1: Performance of Success Predictor and Domain Ranker in identifying the most suitable target domains under domain transfer (DT) and two domain adaptation approaches (SDA and mSDAR).

We can split the data into train and test sets randomly. However, to make sure that the target domain for which we want to select the best source domain has never been seen by the model as the target domain, each time we use one of the 11 domains as the target domain in the test data. Hence, we have 100 training and 10 test samples in each split. For Domain Ranker, we use $F^{S_1 T} \cup F^{S_2 T}$ as the feature set and use the same train-test split as for Success Predictor. This leaves us with 450 training and 45 test samples in each of the 11 splits.

**Success Adaptation Prediction**. Table 1 presents the performance metrics achieved for all target domains. Note that there is a wide variation in success prediction among the target domains. While the Success Predictor achieves good performance on Apple, AskUbuntu and Unix target domains, it performs poorly on PAWS and Quora datasets. This might be due to the difficulty of learning in these domains (Zhang et al., 2019), which is not captured by the descriptive and cross-domain metrics that we use.

**Order Prediction**. Rank correlation coefficients such as Kendall's τ are a common metric to measure the degree of similarity between two rankings. However, here we are more interested in finding out whether we have correctly identified the most relevant source domains. Hence, we report the percentages of top $N$ domains we have identified correctly for $N = 1, 3, 5$. We also report a stricter metric, Correct Rank Percentage (CRP), which equals the percentage of the source domains that have been predicted with the same order as the true ordering. For example, for Stats as the target domain, the true ordering of other domains using SDA is [StackOverflow, AskUbuntu, Apple, Unix, MRPC, SuperUser, SICK, Math, PAWS, Quora]. The Success Predictor predicts the ordering of the source domains as [StackOverflow, Math, Apple, SuperUser, Unix, AskUbuntu, SICK, MRPC, PAWS, Quora]. In this case, CRP=0.5 since 5 out of the 10 domains have the same order in the predicted and true orderings. Also, Top1=1 since the domain with highest predicted order, StackOverflow, has also the highest order in the true ordering. Similarly, Top3=0.67 since only 2 out of the 3 highest ordered domains in true ordering exist in the top 3 of predicted ordering.

**In-Domain and Cross-Domain Performance**

Table 2 shows in-domain and cross-domain performance with DT and DA using absolute F1

| Domain | In-domain average F1 | Domain transfer | | Domain adaptation (SDA) | | Domain adaptation (mSDAR) | |
|---|---|---|---|---|---|---|---|
| | | Average F1 | # of transfer successes | Average F1 | # of adaptation successes | Average F1 | # of adaptation successes |
| Apple | 0.87 | 0.55 | 5 | 0.55 | 4 | 0.56 | 4 |
| AskUbuntu | 0.91 | 0.53 | 4 | 0.55 | 4 | 0.57 | 4 |
| Math | 0.60 | 0.53 | 8 | 0.37 | 4 | 0.51 | 8 |
| MRPC | 0.76 | 0.51 | 5 | 0.66 | 8 | 0.65 | 7 |
| PAWS | 0.26 | 0.49 | 10 | 0.56 | 10 | 0.54 | 10 |
| Quora | 0.78 | 0.50 | 0 | 0.52 | 0 | 0.46 | 0 |
| SICK | 0.59 | 0.47 | 8 | 0.44 | 7 | 0.44 | 7 |
| StackOverflow | 0.94 | 0.49 | 3 | 0.55 | 2 | 0.57 | 3 |
| Stats | 0.82 | 0.43 | 2 | 0.49 | 2 | 0.53 | 4 |
| SuperUser | 0.91 | 0.57 | 5 | 0.56 | 4 | 0.58 | 3 |
| Unix | 0.88 | 0.56 | 4 | 0.55 | 4 | 0.59 | 4 |

Table 2. Average absolute F1 scores for in-domain and cross-domain performance with domain transfer (DT) and domain adaptation (DA). "In-domain" refers to a model trained and evaluated on the same domain, specified in the first column. DT and DA are results for a model *trained* on other domains (source) and *evaluated* on the domain in the first column (target). For DT and DA, "# of transfer/adaptation successes" is the number of source domains (out of 10) where a model evaluated on target performed at least at 80% of in-domain performance.

scores. Comparing columns "Average F1" with "In-domain average F1", DT and DA performance for most domains is lower than in-domain performance. The exception is PAWS, where DA delivers over twice the performance of in-domain training.

Additionally, for most domains DT and DA resulted in some successes and some failures (mostly between 3 and 8 successes for mSDAR). The exception was, again, PAWS, where all source domains succeeded in DT/DA, and Quora, where none succeeded.

## 7 Discussion

**Adaptation Success Factors**. Comparing CPR, Top1, Top3 and Top5 between Success Predictor and Domain Ranker for all DT and DA methods, we see that in general, Domain Ranker does a better job finding the orderings of candidate source domains. For Success Predictor, the features with highest importance are KL-divergence, $f_8^{ST}$, Target Ratio, $f_2^{ST}$, and data size of the target domain $f_4^{ST}$. For Domain Ranker, average example lengths, $f_5^{S_1T}$, $f_5^{S_2T}$, Rényi divergences $f_7^{S_1T}$, $f_7^{S_2T}$ and perplexities $f_9^{S_1T}$, $f_9^{S_2T}$ in both source domains are the most informative features.

**Adaptation Success by Dataset**. While the Success Predictor performed reasonably well on most domains, its performance on PAWS and Quora datasets was miserable. We attribute this to the lack of domain similarity features that would reflect the complexities of these datasets, and note this for future work. The PAWS result can be explained by a representation and training we used (USE and dense neural network), which is different from bag-of-words and BERT used by PAWS authors (Zhang et al., 2019). For Quora, in-domain performance is in line with previous research (Wang et al., 2017, Tomar et al., 2017), and aggregate DT/DA results were similar to other datasets such as Stats.

## 8 Conclusion and Future Work

We studied the problem of selecting the most relevant labeled datasets from a pool of candidates to be used as a source domain in a transfer learning setup with a specific unlabeled target domain. The experiments focused on the text similarity task and autoencoder approaches to DA. Note that the proposed process can be extended to other NLP tasks and other unsupervised DA approaches as well. We used descriptive domain information and cross-domain similarity metrics as predictive features to model the success of DT and DA, and to rank source domains based on their relevancy.

In future work, we intend to study source selection in *multi-source* domain adaptation setup, using multiple source domains for DT/DA. Identifying additional adaptation success factors that could better predict the success of DT/DA for complex domains such as PAWS and Quora, and learning the success threshold (here, we fixed it at 80%) are other avenues to investigate. Other possibilities include experimenting with various text representations (such as bag-of-words) and models (e.g., Transformer-based).